\documentclass[11pt, a4paper]{article}

\usepackage[a4paper, margin=2.5cm]{geometry}
\usepackage[english]{babel}
\usepackage{amsmath, amssymb}
\usepackage{booktabs}
\usepackage{tabularx}
\usepackage{multirow}
\usepackage{graphicx}
\usepackage{xcolor}
\usepackage{float}
\usepackage{enumitem}
\usepackage{makecell}
\usepackage[table]{xcolor}
\usepackage[colorlinks=true, linkcolor=blue, citecolor=blue, urlcolor=blue]{hyperref}

\newcolumntype{Y}{>{\raggedright\arraybackslash}X}
\newcolumntype{C}{>{\centering\arraybackslash}X}

\title{\textbf{Conditional Diffusion-Based Generation of T2-Weighted Prostate MRI for Clinically Significant Prostate Cancer Classification}}

\author{
  Xavier Rafael-Palou$^{1}$, Alejandro Vergara-Richart$^{1,2}$, Ignacio Iborra Roncales$^{1}$, \\
  Carmen Prieto-de-la-Lastra$^{1}$, Andreu Antolin-Redondo$^{3,4}$, Juan Morote$^{4}$, \\
  Ana Jimenez-Pastor$^{1}$
}

\date{}

\begin{document}

\maketitle

\vspace{-2em}
\begin{center}
  \small
  $^{1}$Quantitative Imaging Biomarkers in Medicine (Quibim S.L.), Valencia, Spain\\
  $^{2}$Universitat Politècnica de València, València, Spain\\
  $^{3}$Department of Radiology, Institut de Diagnòstic per la Imatge (IDI), Hospital Universitari Vall d'Hebron, Barcelona, Spain\\
  $^{4}$Department of Surgery, Universitat Autònoma de Barcelona, Bellaterra, Spain
\end{center}

\vspace{1.5em}

\begin{abstract}
\noindent \textbf{Objectives:} Scarce and unrepresentative training data limit the performance of prostate MRI prediction models. Synthetic T2-weighted (T2W) MRI may augment training data containing clinically significant prostate cancer (csPCa) and transition zone (TZ) lesions. We developed a conditional generative model for T2W prostate MRI and evaluated its fidelity and utility for csPCa classification under simulated data constraints.

\smallskip
\noindent \textbf{Materials and methods:} In this retrospective, single-center, ethics-committee-approved study, we developed a generative model extending MAISI-v2 with csPCa class, zonal, and lesion-mask-guided conditioning, incorporating a relative advantage weighted loss, fine-tuned on 1,303 patients (1,654 lesions). Fidelity was assessed by incremental ablation. Downstream utility was evaluated with T2W-based csPCa classifiers across five simulated training-data scenarios (full data, low-data regime, class imbalance, TZ lesion underrepresentation, and combined), each testing 18 synthetic-data configurations varying dataset size, class proportion and case selection. The best configuration by mean validation area under the curve (AUC) across five seeds was compared with a real-data-only baseline on the test set (paired t-tests, Benjamini--Hochberg correction). PI-CAI was used for external fidelity assessment and cross-institutional augmentation.

\smallskip
\noindent \textbf{Results:} The proposed model reduced Fréchet inception distance (FID) from 13.28 (pretrained MAISI-v2) to $2.04 \pm 0.06$ internally (PI-CAI: $3.83 \pm 0.51$). Synthetic augmentation significantly increased AUC in all five scenarios (+0.027 to +0.088; adjusted $p = 0.016$--$0.043$). Cross-institutional augmentation of PI-CAI yielded a non-significant gain (+0.013; $p = 0.397$).

\smallskip
\noindent \textbf{Conclusions:} Our model improved T2W MRI fidelity over pretrained MAISI-v2, and synthetic images improved T2W-only csPCa classification under simulated data constraints, although benefit did not transfer significantly across institutions.
\end{abstract}


\section*{Key Points}
\begin{itemize}[leftmargin=*]
    \item \textbf{Question:} Whether a conditional generative model for T2-weighted prostate MRI improves image fidelity and clinically significant prostate cancer classification when training data are scarce or imbalanced.
    \item \textbf{Findings:} The model reduced Fréchet inception distance from 13.28 to 2.04, and synthetic augmentation increased area under the curve by 0.027--0.088 across five simulated scenarios.
    \item \textbf{Relevance Statement:} Our conditional generative model produced high-fidelity T2-weighted images, and these synthetic images improved clinically significant prostate cancer classification when real training data were scarce or imbalanced, including underrepresented transition zone lesions. Cross-institutional benefit requires site-specific verification.
\end{itemize}

\section*{List of Abbreviations}
\begin{table}[H]
\centering
\small
\begin{tabularx}{\textwidth}{lX}
\toprule
\textbf{Abbreviation} & \textbf{Definition} \\
\midrule
ADC & Apparent diffusion coefficient \\
AUC & Area under the receiver operating characteristic curve \\
BH & Benjamini--Hochberg \\
CNet & ControlNet \\
csPCa & Clinically significant prostate cancer \\
CV & Cross-validation \\
DWI & Diffusion-weighted imaging \\
EMA & Exponential moving average \\
FID & Fréchet inception distance \\
FoV & Field of view \\
GG & Grade Group \\
iPCa & Insignificant prostate cancer \\
ISUP & International Society of Urological Pathology \\
KID & Kernel inception distance \\
LDM & Latent diffusion model \\
MAISI & Medical AI for Synthetic Imaging \\
mpMRI & Multiparametric MRI \\
PCa & Prostate cancer \\
PI-CAI & Prostate Imaging: Cancer AI \\
PZ & Peripheral zone \\
R & Real data only \\
R+S & Real plus synthetic data \\
relLoss & Relative advantage loss \\
relWLoss & Relative advantage weighted loss \\
T2W & T2-weighted \\
TR & Real training set \\
TR-inv & Inverse of the real training-set class ratio \\
TR-ratio & Real training-set class ratio \\
TZ & Transition zone (central plus transition zone) \\
VAE & Variational autoencoder \\
wLoss & Region-weighted loss \\
\bottomrule
\end{tabularx}
\end{table}

\section{Introduction}
Multiparametric MRI is the recommended first-line test for suspected prostate cancer (PCa), aiming to detect clinically significant PCa (csPCa; ISUP Grade Group [GG] $\ge$ 2), which usually warrants radical treatment, while avoiding unnecessary biopsies of benign findings and overdiagnosis of insignificant PCa (iPCa; GG1), typically managed with active surveillance \cite{epstein2016, cornford2026}. However, variable image quality and limited inter-reader agreement, despite PI-RADS standardisation \cite{turkbey2019, greer2019}, reduce diagnostic consistency \cite{stabile2020}, motivating AI-based support \cite{antolin2026}.

Deep learning models using T2-weighted (T2W), diffusion-weighted imaging (DWI), and apparent diffusion coefficient (ADC) maps show promising but heterogeneous csPCa detection, with generalisability constrained by small cohorts \cite{roest2022}, class imbalance \cite{saha2024}, and underrepresentation of less prevalent lesions, such as transition zone (TZ) cancers \cite{mcneal1988}.

Synthetic images can mitigate these constraints by expanding training cohorts, easing data-sharing \cite{ktena2024}, and improving robustness to distribution shift \cite{brugnara2024}. Prostate MRI synthesis first relied on generative adversarial networks \cite{xu2023, yang2020, fernandez2021}; diffusion models then offered more stable training and better distribution coverage \cite{dhariwal2021}, latent diffusion \cite{rombach2022} and rectified flow \cite{liu2023} improved efficiency, and conditioning enabled targeted oversampling \cite{kazerouni2023}. Generalist 3D frameworks such as 3D MedDiffusion \cite{wang2025} and MAISI-v2 \cite{zhao2026} extend these advances to volumetric, multi-modality synthesis.

Few diffusion-based studies have synthesised prostate MRI, mostly T2W images for prostate and lesion segmentation \cite{bashkanov2024, giardina2025}. Synthetic T2W augmentation has improved csPCa classification, albeit only in an in-house cohort \cite{grabke2025a}; and multi-sequence synthesis (T2W, ADC, and high-b-value DWI) improved cross-institutional generalization over T2W-only \cite{grabke2025b}. Multi-sequence synthesis, however, requires large, harmonized cohorts, whereas T2W is routinely acquired and widely available, although insufficient alone for comprehensive csPCa classification \cite{shinmoto2015}. High-fidelity synthetic T2W could still enlarge and diversify training cohorts and strengthen T2W encoders for multimodal models beyond csPCa classification \cite{lowry2026}, yet its downstream utility remains insufficiently investigated, particularly under class imbalance and TZ lesion underrepresentation.

In this study, we hypothesized that a conditional generative model would improve T2W prostate MRI fidelity over MAISI-v2 \cite{zhao2026}, and that its synthetic images would improve T2W-only csPCa classification under simulated training-data constraints. We therefore extended MAISI-v2 to condition generation on prostate zone and lesion masks and csPCa status, and compared classifiers trained on real data alone or augmented with synthetic images across five scenarios: full training data, a low-data regime, severe class imbalance, TZ lesion underrepresentation, and underrepresentation of csPCa+ lesions in the TZ. The PI-CAI dataset \cite{saha2024} was used to assess external fidelity and cross-institutional transfer.

\section{Methods}

\subsection{Study design and dataset}
This retrospective study was approved by the Ethics Committee of Hospital Universitari Vall d'Hebron (Barcelona, Spain; PR(AG)02/2021), which waived informed consent, and followed the Declaration of Helsinki.

We used the previously reported cohort of Antolin et al. \cite{antolin2026}: 1,395 men with suspected PCa consecutively selected at a single center (November 2015--November 2022), imaged at 1.5T or 3T (MAGNETOM Symphony/Trio, Siemens Healthineers) and scored with PI-RADS v2 or v2.1 \cite{turkbey2019}. Men with PI-RADS $\ge 3$ underwent targeted plus 12-core systematic biopsy; others, systematic biopsy only (eligibility criteria and protocols in \cite{antolin2026}). After excluding 92 patients with preprocessing failure, 1,303 (93.4\%; 1,654 lesions) remained; no clinical exclusions were applied. Lesions with biopsy GG $\ge 2$ were labeled csPCa+ and the rest csPCa$-$ (iPCa or benign); patients were csPCa+ if any lesion was.

Bi-parametric MRI was collected and pseudonymized via the QP-Care\textregistered{} platform, and QP-Prostate\textregistered{} automatically segmented, in T2W space, the prostate zones (peripheral zone [PZ], and central plus transition zone, hereafter TZ) and suspicious csPCa lesions. An abdominal radiologist (5 years of experience) refined the lesion masks and delineated lesions missed by the software (GG $<$ 2).

For generative model training, each case comprised one T2W volume paired with a mask of the prostate zones and one lesion; lesions spanning both zones were assigned to the zone of greatest overlap. As patients contributed several cases (multiple lesions or repeat MRI examinations; Supplementary Material), two datasets were derived (Table \ref{tab:cohort}). The patient-level dataset (one case per patient; $n = 1,303$) was used to train the stage without mask conditioning, avoiding overrepresentation of these patients. A case with a csPCa+ lesion was selected when available, so that it carried the patient-level label (ties resolved randomly with a fixed seed to ensure reproducibility). The lesion-level dataset (one case per lesion; $n = 1,654$) was used to train the mask-conditioned stage. Both datasets were split at patient level into training, validation, and test sets (80/10/10), so that all cases from the same patient remained in one set to prevent data leakage. This yielded 1,047/126/130 cases in the patient-level dataset and 1,335/159/160 in the lesion-level dataset.

\begin{table}[H]
\centering
\label{tab:cohort}
\vspace{0.5em}
\scriptsize
\begin{tabularx}{\textwidth}{l*{6}{C}}
\toprule
 & \multicolumn{2}{c}{\textbf{Train}} & \multicolumn{2}{c}{\textbf{Validation}} & \multicolumn{2}{c}{\textbf{Test}} \\
\cmidrule(lr){2-3} \cmidrule(lr){4-5} \cmidrule(lr){6-7}
 & \textbf{Lesion level} & \textbf{Patient level} & \textbf{Lesion level} & \textbf{Patient level} & \textbf{Lesion level} & \textbf{Patient level} \\
\midrule
Cases & 1,335 & 1,047 & 159 & 126 & 160 & 130 \\
csPCa$-$ (csPCa+) & 855 (480) & 609 (438) & 112 (47) & 84 (42) & 106 (54) & 81 (49) \\
Lesion PZ (TZ) & 877 (458) & 683 (364) & 110 (49) & 90 (36) & 100 (60) & 83 (47) \\
Prostate vol. (mL) & 53.0 (40.0, 73.0) & 53.0 (40.0, 74.0) & 60.0 (42.0, 87.0) & 59.5 (40.0, 84.8) & 52.0 (35.0, 78.3) & 52.5 (37.0, 79.0) \\
Spacing XY (mm) & 0.625 (0.56, 0.63) & 0.625 (0.56, 0.63) & 0.625 (0.56, 0.63) & 0.625 (0.56, 0.63) & 0.625 (0.56, 0.63) & 0.625 (0.56, 0.63) \\
Spacing Z (mm) & 3.0 (3.0, 3.0) & 3.0 (3.0, 3.0) & 3.0 (3.0, 3.0) & 3.0 (3.0, 3.0) & 3.0 (3.0, 3.0) & 3.0 (3.0, 3.0) \\
Shape XY & 256 (256, 320) & 316 (256, 320) & 256 (256, 320) & 256 (256, 320) & 316 (256, 320) & 316 (256, 320) \\
Shape Z & 40 (40, 40) & 40 (40, 40) & 40 (40, 40) & 40 (40, 40) & 40 (40, 40) & 40 (40, 40) \\
FoV XY (mm) & 177.8 (160, 180) & 177.8 (160, 180) & 160.0 (160, 177.8) & 160.0 (160, 180) & 177.8 (160, 180) & 177.8 (160, 180) \\
FoV Z (mm) & 120 (120, 120) & 120 (120, 120) & 120 (120, 120) & 120 (120, 120) & 120 (120, 120) & 120 (120, 120) \\
\bottomrule
\end{tabularx}
\caption{Cohort characteristics of the training, validation, and test partitions, reported at the lesion level and patient level. Continuous variables are reported as median (interquartile range, 25th--75th percentiles). csPCa$-$ (csPCa+), cases without (with) clinically significant prostate cancer; Lesion PZ (TZ), cases with the lesion in the peripheral (transition) zone; FoV, field of view.}
\end{table}

\begin{figure}[H]
\centering
\includegraphics{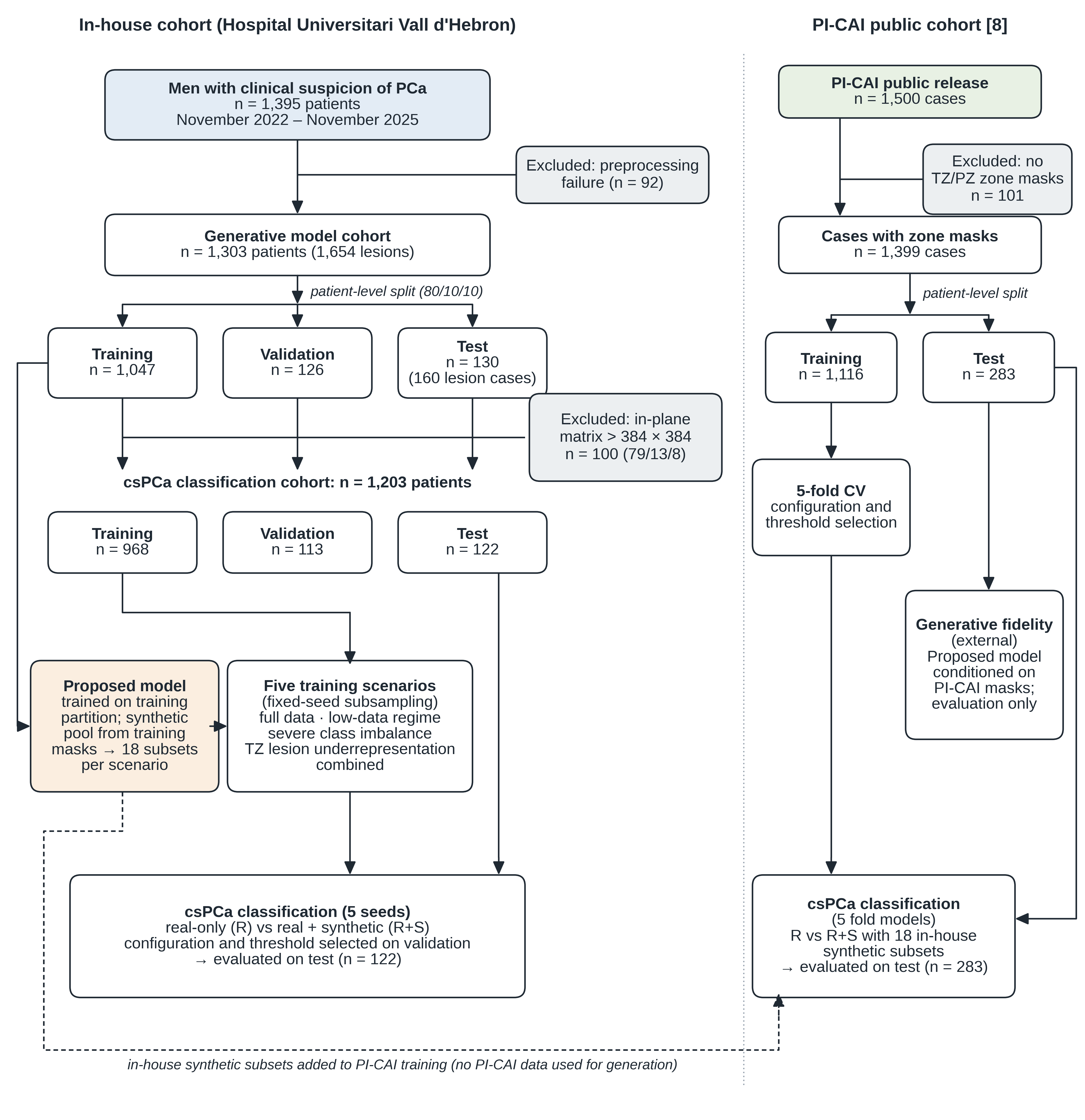}
\caption{Study flowchart. Left, in-house cohort: exclusions, patient-level partitions for generative model development and csPCa classification, and the five training scenarios augmented with the proposed generative model. Right, PI-CAI cohort: patient-level split, 5-fold cross-validation for configuration and threshold selection, external generative fidelity evaluation, and cross-institutional classification with in-house synthetic subsets. PCa, prostate cancer; csPCa, clinically significant PCa; PZ, peripheral zone; TZ, transition zone; CV, cross-validation; R, real-only; R+S, real plus synthetic.}
\label{fig:flowchart}
\end{figure}

The public PI-CAI dataset \cite{saha2024} served as an independent cohort for external and cross-institutional evaluation. Of its 1,500 T2W volumes, only 1,399 with prostate zone masks (generated by QP-Prostate\textregistered{}) were included, with csPCa labels as provided. Ethics approval was obtained by the original data providers, and data were used under the dataset's terms of use. Volumes were randomly split at patient level into training ($n = 1,116$; 80\%) and test ($n = 283$; 20\%) sets. The training set was used to develop classifiers combining real PI-CAI with synthetic images, with 5-fold cross-validation for model selection; the held-out test set was used to evaluate these classifiers and synthetic image fidelity. Figure \ref{fig:flowchart} summarizes the workflow.

\subsection{Generative image method}
\begin{figure}[H]
\centering
\includegraphics[width=0.7\linewidth]{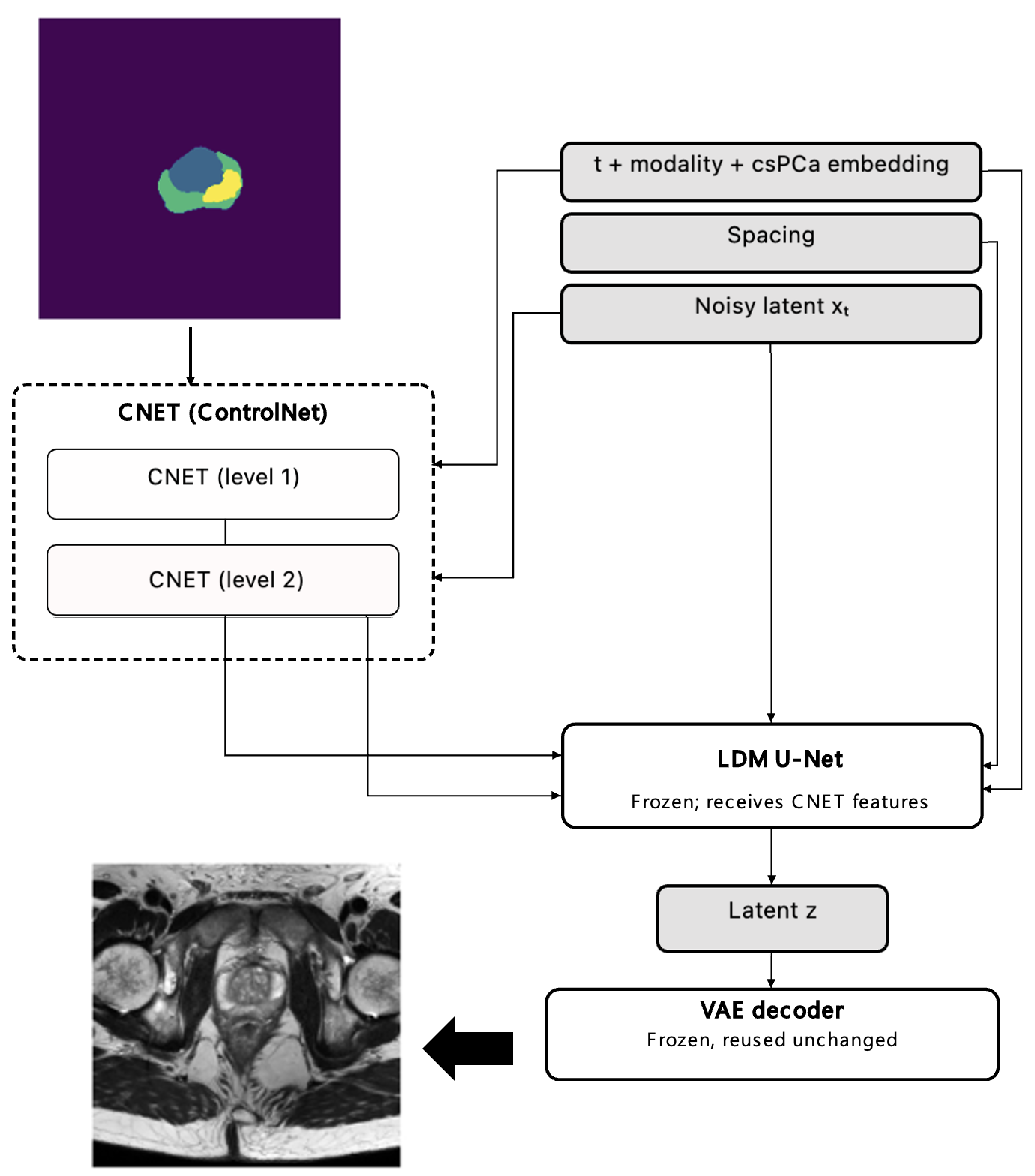}
\caption{Overview of the proposed generative model, comprising three networks: variational autoencoder (VAE), latent diffusion model (LDM) and ControlNet (CNET). The ControlNet receives the multi-class zone and lesion mask, the noisy latent ($x_t$), and the timestep ($t$), modality and csPCa embedding, and injects its features into the frozen LDM U-Net, which also receives the voxel spacing; the frozen VAE decoder maps the generated latent ($z$) to the synthetic T2-weighted image. csPCa, clinically significant prostate cancer.}
\label{fig:architecture}
\end{figure}

The proposed method (Figure \ref{fig:architecture}) builds on the open-source MAISI-v2 rflow-mr release \cite{zhao2026}, a 3D latent diffusion framework comprising a variational autoencoder (VAE) and a latent diffusion model (LDM), both pretrained on large (CT and MRI) cohorts spanning diverse anatomical regions. The VAE encodes 3D volumes into a lower-dimensional latent space and decodes latents back to images. The LDM operates in this latent space under a rectified-flow formulation \cite{liu2023}: during training, a latent is linearly interpolated with Gaussian noise at a random timestep, and the network learns to predict the velocity field between noise and data. At inference, Gaussian noise is iteratively updated by integrating the learned velocity over a fixed number of steps, and the resulting latent is decoded by the VAE.

\subsection{Adaptations for prostate MRI synthesis}
Four additions were introduced:
\begin{enumerate}[leftmargin=*]
    \item \textbf{csPCa class conditioning.} The binary csPCa label was injected into the LDM through the framework's conditioning embedding, encouraging class-specific lesion appearance and enabling class control of synthetic data.
    \item \textbf{Exponential moving average (EMA) \cite{song2020}.} An exponentially averaged copy of the LDM weights was maintained and updated during training, to reduce parameter noise and improve inference.
    \item \textbf{ControlNet \cite{zhang2023}.} A trainable copy of the LDM encoder was added, taking a multi-class zone and lesion mask as input (with the noised latent, timestep, image modality type and csPCa status) and injecting its outputs into the frozen LDM, conditioning synthesis on zonal anatomy and lesion location.
    \item \textbf{Relative advantage weighted loss (relWLoss).} A relative advantage loss ($L_{\text{rel}}$) enforced the use of mask conditioning by penalizing the conditioned prediction ($v_{\text{ctrl}}$, from the ControlNet and frozen LDM) whenever its velocity error was not at least a margin $m$ below that of the unconditioned prediction ($v_{\text{ldm}}$, from the frozen LDM alone). It was combined, weighted by $\alpha$, with the framework's region-weighted loss ($L_w$) (Eqs. \ref{eq:1}--\ref{eq:3}).
\end{enumerate}

\begin{equation}
L_{\text{rel}} = \max\left(0, e_{\text{ctrl}} - e_{\text{ldm}} + m\right), \quad e_{\text{ctrl}} = \|v_{\text{ctrl}} - v^*\|_1, \quad e_{\text{ldm}} = \|v_{\text{ldm}} - v^*\|_1
\label{eq:1}
\end{equation}
\begin{equation}
L_w = \text{mean}\left(|v_{\text{ctrl}} - v^*| \odot w \cdot M\right)
\label{eq:2}
\end{equation}
\begin{equation}
L_{\text{relW}} = L_w + \alpha \cdot L_{\text{rel}}
\label{eq:3}
\end{equation}

\noindent Where $v^*$ is the target velocity, $M$ is the mask, $w$ is the region weight, $\odot$ denotes element-wise multiplication, and $\alpha$ balances both terms.

\subsection{Training}
Training was performed in two stages (hyperparameters in the Supplementary Material).

\textbf{Stage 1.} The LDM was initialized with MAISI-v2 rflow-mr weights and fine-tuned without spatial conditioning on the patient-level dataset ($n = 1,047$). Volumes were foreground-cropped, resized to $256 \times 256 \times 128$, clipped at 95.5th percentiles, normalized to $[0, 1]$, and encoded once with the frozen VAE. LDM inputs were the csPCa label, image modality type, voxel spacing, and timestep; optimization used rectified flow with L1 loss and EMA.

\textbf{Stage 2.} The frozen LDM and ControlNet were initialized from Stage-1 EMA weights. Only the ControlNet was fine-tuned, on lesion-level cases ($n = 1,335$) with relWLoss. Masks were cropped, resized with nearest-neighbor interpolation, and one-hot encoded \cite{chen2023}.

\subsection{Downstream task: csPCa classification}
Augmentation utility was evaluated on patient-level csPCa classification from T2W alone, acknowledging its known underperformance relative to mpMRI \cite{shinmoto2015}. To isolate the gain attributable to synthetic data, matched classifier pairs were trained on real data only (R) and on real plus synthetic cases (R+S).

Real data comprised one case per patient and its T2W volume was resampled to the training-set median voxel spacing ($0.625 \times 0.625 \times 3.0$ mm) and to a $256 \times 256 \times 128$ matrix (median dimensions rounded to multiples of 128, satisfying the MAISI-v2 divisibility \cite{zhao2026}), matching the synthetic volume geometry. Cases with an in-plane matrix exceeding $384 \times 384$ voxels (92.5th training-set percentile) were excluded to limit interpolation-related real-synthetic discrepancies. The downstream dataset comprised 968/113/122 cases for training, validation and test, matching the generative model's partitions to prevent leakage.

The synthetic pool was generated conditioned only on downstream training-partition masks and csPCa labels, so no validation or test information contributed to it; validation and test sets contained real data only.

csPCa classifiers used a frozen, pretrained SwinTransformer backbone \cite{tang2022} with a lightweight head (two dense layers, dropout, batch normalization), trained under identical settings (Supplementary Material) to avoid fine-tuning bias, with no augmentation beyond the synthetic cases.

\subsection{Evaluation framework}
We evaluated 18 synthetic configurations in each of five simulated training-data scenarios.

\subsubsection{Evaluation scenarios}
The downstream training partition was subsampled (fixed seed) to simulate: full training data, a low-data regime, severe class imbalance, TZ lesion underrepresentation, and underrepresentation of csPCa+ TZ lesions (Table \ref{tab:scenarios}). Validation and test sets were not subsampled.

\begin{table}[H]
\centering
\label{tab:scenarios}
\vspace{0.5em}
\scriptsize
\begin{tabularx}{\textwidth}{>{\raggedright\arraybackslash}p{2.0cm} >{\raggedright\arraybackslash}X *{6}{c}}
\toprule
 & & \multicolumn{6}{c}{\textbf{Scenario training set (real data)}} \\
\cmidrule(lr){3-8}
\textbf{Scenario} & \textbf{Constraint applied} & \textbf{Total} & \textbf{csPCa+} & \textbf{csPCa$-$} & \makecell[tl]{TZ \\Lesions} & \makecell[tl]{PZ \\Lesions} & \makecell[tl]{csPCa+ \\ \& TZ} \\
\midrule
\arrayrulecolor{lightgray}
Full data & None. All downstream training set & 968 & 394 & 574 & 345 & 623 & 147 \\
Low data & 30\% of downstream training set & 290 & 118 & 172 & 103 & 187 & 49 \\
csPCa+ imbal. & 30\% of csPCa+ lesions from downstream train & 693 & 119 & 574 & 247 & 446 & 49 \\
TZ imbal. & 30\% of TZ lesions from downstream train & 727 & 296 & 431 & 104 & 623 & 49 \\
csPCa+/TZ imbal. & 30\% of csPCa+ TZ lesions from downstream train & 866 & 292 & 574 & 243 & 623 & 45 \\
\arrayrulecolor{black} 
\bottomrule
\end{tabularx}
\caption{Simulated training-data scenarios and characteristics of their real training sets (patient level). csPCa, clinically significant prostate cancer; PZ, peripheral zone; TZ, transition zone.}
\end{table}

\subsubsection{Data generation criteria}
For each scenario, 18 synthetic subsets were drawn from the pool differing in the number of synthetic images, class proportion and case selection (Table \ref{tab:config}). One csPCa classifier was trained per subset (real + synthetic), alongside a real-only baseline.

\begin{table}[H]
\centering
\label{tab:config}
\vspace{0.5em}
\small
\begin{tabularx}{\textwidth}{l l X}
\toprule
\textbf{Factor} & \textbf{Levels} & \textbf{Description} \\
\midrule
Synthetic size & 10\%, 20\%, 30\% & Proportion of synthetic images relative to the real training set of each scenario. \\
\addlinespace
Class proportion & Balanced, TR-ratio, TR-inv & csPCa+/csPCa$-$ ratio of synthetic subset. \textbf{Balanced}: equal csPCa+/csPCa$-$. \textbf{TR-ratio}: same ratio as real training set. \textbf{TR-inv}: inverse ratio (e.g., 30:70 becomes 70:30) to counterbalance imbalance. \\
\addlinespace
Case selection & Fixed, Bounded & Selection criterion from generated pool. \textbf{Fixed}: first N images without filtering. \textbf{Bounded}: uncertainty-guided, retaining images with predicted csPCa+ probability between 0.40 and 0.60. \\
\bottomrule
\end{tabularx}
\caption{Design of synthetic augmentation configurations (18 configurations; $3 \times 3 \times 2$). csPCa, clinically significant prostate cancer; TR, real training set.}
\end{table}

\subsubsection{Cross-institutional evaluation}
To test cross-institutional transfer, classifiers were trained on the PI-CAI training set alone or augmented with each of the 18 in-house synthetic subsets and evaluated on the PI-CAI test set.

\subsubsection{Evaluation metrics}
Synthetic image fidelity and diversity were assessed with Fréchet Inception Distance (FID) \cite{heusel2017}, Kernel Inception Distance (KID) \cite{binkowski2018}, and generative precision and recall \cite{kynkaanniemi2019}. Each metric compared the test set with an equally sized synthetic set (patient-level for LDM-only and lesion-level for ControlNet-based models). Following a 2.5D approach \cite{pinaya2023}, features were extracted from all axial, coronal, and sagittal slices with a RadImageNet-pretrained ResNet50 \cite{mei2022}, and metrics were computed per plane and averaged. Validation FID guided generative model selection.

csPCa classification was assessed with AUC, sensitivity, specificity, precision, and F1-score, at a threshold fixed at the F1-optimal point on validation. Within each scenario, all 18 configurations and the real-only baseline were trained with the same five seeds differing only in network initialization. The configuration with the highest mean validation AUC was selected for testing, and results are reported as mean $\pm$ SD across seeds. On PI-CAI, the configuration and threshold were selected likewise by 5-fold cross-validation within the training split, and test metrics are reported as mean $\pm$ SD across the fivefold models. For each scenario, per-seed AUCs of the selected configuration and the real-only baseline were compared with two-sided paired t-tests and reported as mean difference with 95\% confidence interval. P-values across the five in-house scenarios were adjusted with the Benjamini--Hochberg (BH) procedure (significance, $p < 0.05$); the cross-institutional comparison, addressing a separate question, is reported unadjusted. Analyses used Python 3.10.4 and statsmodels 0.15.0.

\section{Results}
Of 1,395 in-house patients, 1,303 (1,654 lesions) were included for generative modeling and 1,203 for csPCa classification (Figure \ref{fig:flowchart}; Table \ref{tab:cohort}); PI-CAI comprised 1,399 cases.

\subsection{Generative quality}
\begin{figure}[H]
\centering
\includegraphics[width=1.0\linewidth]{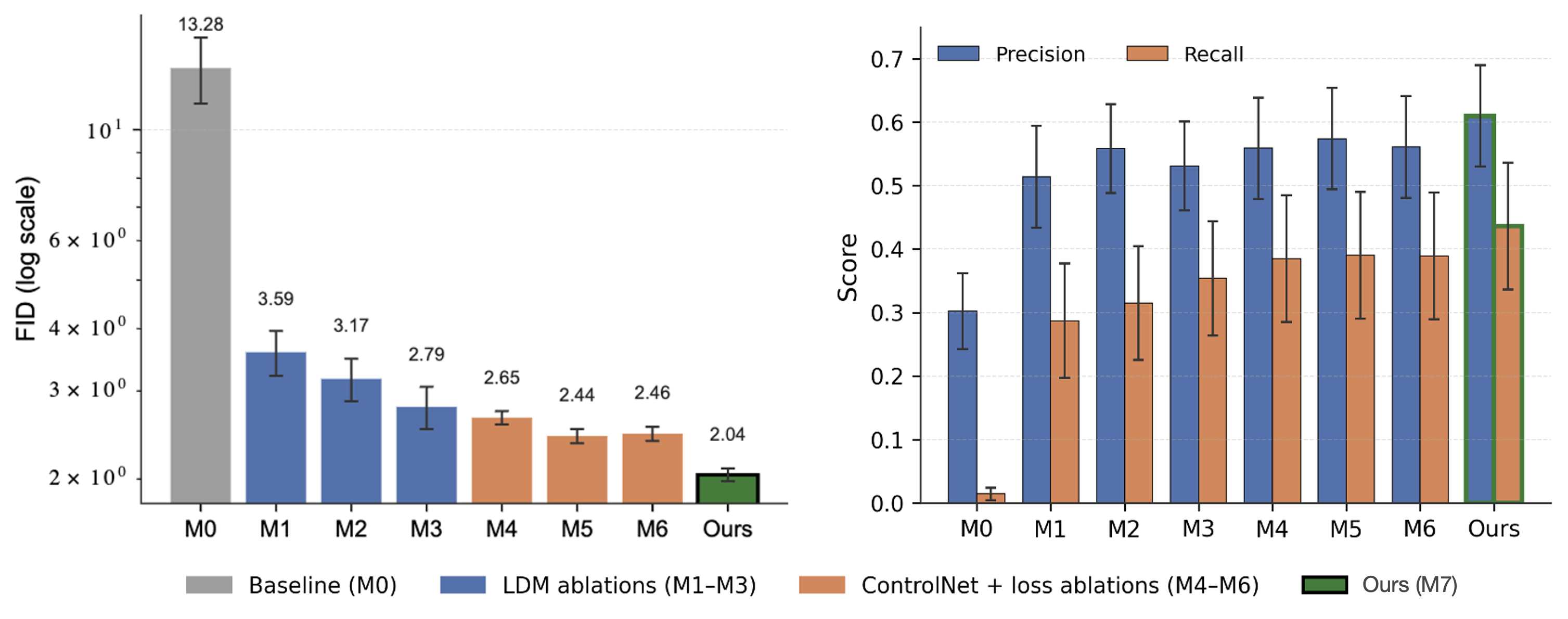}
\caption{Generative ablation results on the in-house test set: Fréchet inception distance (FID; left, log scale; values above bars are means) and precision and recall (right) for the pretrained baseline (M0), latent diffusion model (LDM) ablations (M1--M3), ControlNet and loss ablations (M4--M6), and the proposed model (M7, Ours). Error bars represent SD across the three orthogonal planes.}
\label{fig:generative_ablation}
\end{figure}

\begin{figure}[H]
\centering
\includegraphics[width=1.0\linewidth]{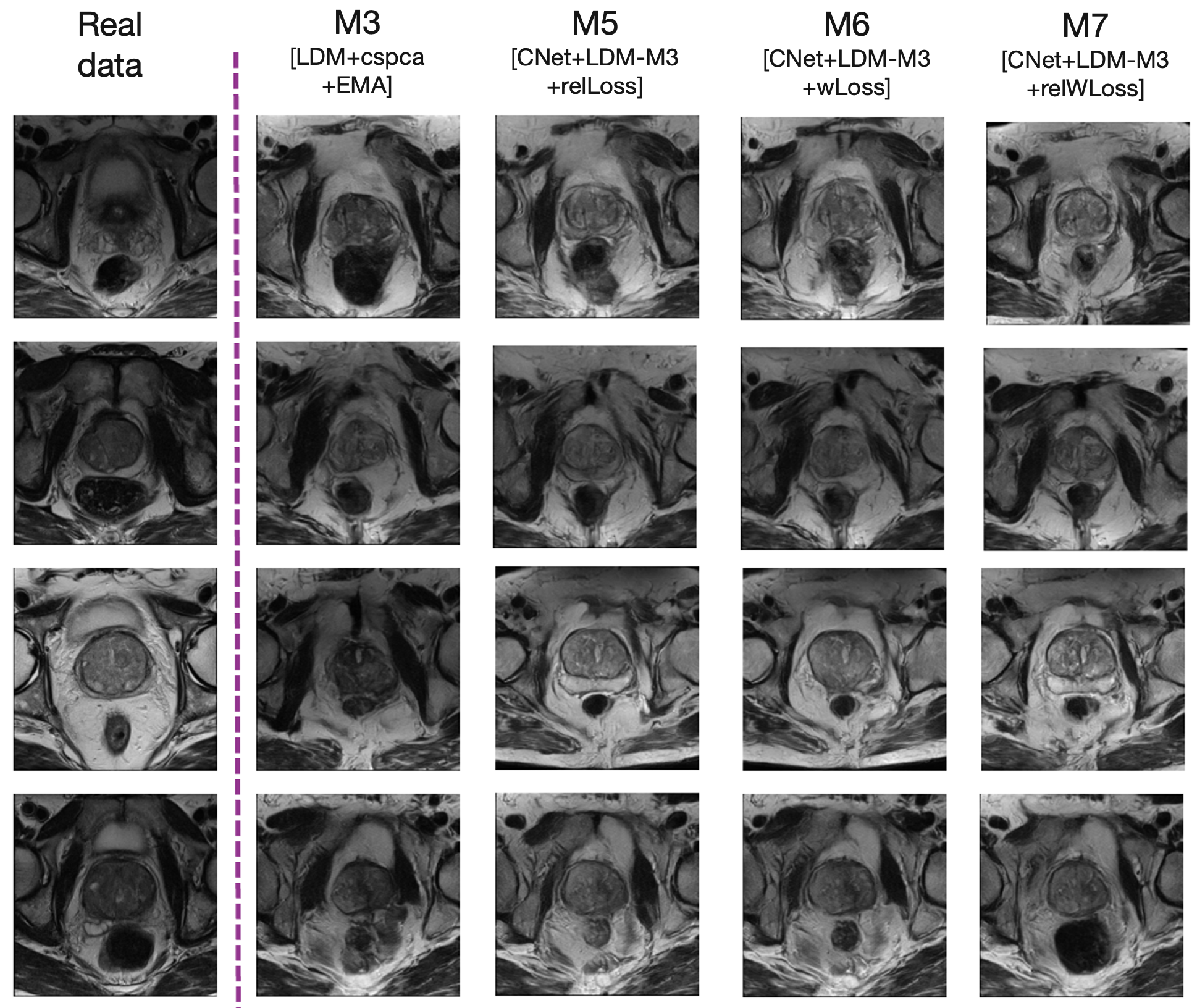}
\caption{Representative axial T2-weighted slices of four cases (rows): real image (first column) and synthetic images from M3 (best LDM without mask conditioning; second column), M5 (ControlNet with relative advantage loss; third column), M6 (ControlNet with region-weighted loss; fourth column), and M7 (proposed model, ControlNet with relative advantage weighted loss; last column).}
\label{fig:examples}
\end{figure}

Fine-tuning produced the largest improvement over pretrained MAISI-v2 (FID 13.28 to 3.59; M0 to M1), and class conditioning and EMA further reduced FID to 2.79 (M3) (Table \ref{tab:ablation}; Figure \ref{fig:generative_ablation}). Recall increased steadily from M1 to M3 (0.29 to 0.35), whereas precision peaked with class conditioning (M2, 0.56) and decreased slightly with EMA (M3, 0.53). Among lesion-level ControlNet variants, FID decreased from 2.65 (M4) to 2.44 with relLoss (M5), 2.47 with wLoss (M6), and 2.04 with relWLoss (M7), whose reduction ($-0.61$) exceeded the sum of its individual terms ($-0.21$ and $-0.19$); M7 achieved the highest precision (0.61) and recall (0.44). Figure \ref{fig:examples} shows representative examples.

\begin{table}[H]
\centering
\label{tab:ablation}
\vspace{0.5em}
\scriptsize
\begin{tabularx}{\textwidth}{l >{\raggedright\arraybackslash}X c c c c c}
\toprule
\makecell[tl]{ID} & \makecell[tl]{Main \\ components} & \makecell[t]{N Test} & \makecell[t]{FID} & \makecell[t]{KID} & \makecell[t]{Prec} & \makecell[t]{Rec} \\
\midrule
\makecell[tl]{M0 \\ (MAISI-v2)} & LDM & 130 & $13.28 \pm 2.01$ & $0.0093 \pm 4.7\times 10^{-4}$ & $0.302 \pm 0.06$ & $0.014 \pm 0.01$ \\
M1 & LDM & 130 & $3.587 \pm 0.37$ & $0.0027 \pm 5.2\times 10^{-4}$ & $0.514 \pm 0.08$ & $0.287 \pm 0.09$ \\
M2 & LDM + csPCa & 130 & $3.173 \pm 0.31$ & $0.0023 \pm 4.4\times 10^{-4}$ & $0.558 \pm 0.07$ & $0.315 \pm 0.09$ \\
M3 & LDM + csPCa + EMA & 130 & $2.789 \pm 0.27$ & $0.0019 \pm 3.8\times 10^{-4}$ & $0.531 \pm 0.07$ & $0.354 \pm 0.09$ \\
M4 & CNet + LDM (M3) & 160 & $2.653 \pm 0.08$ & $0.0019 \pm 9.4\times 10^{-4}$ & $0.559 \pm 0.08$ & $0.385 \pm 0.10$ \\
M5 & CNet + LDM (M3) + relLoss & 160 & $2.439 \pm 0.08$ & $0.0017 \pm 8.1\times 10^{-4}$ & $0.574 \pm 0.08$ & $0.390 \pm 0.10$ \\
M6 & CNet + LDM (M3) + wLoss & 160 & $2.465 \pm 0.08$ & $0.0017 \pm 9.4\times 10^{-4}$ & $0.561 \pm 0.08$ & $0.389 \pm 0.10$ \\
\makecell[tl]{M7 \\ (Ours)} & CNet + LDM (M3) + relWLoss & \textbf{160} & $\mathbf{2.039 \pm 0.06}$ & $\mathbf{0.0014 \pm 8.1\times 10^{-4}}$ & $\mathbf{0.610 \pm 0.08}$ & $\mathbf{0.436 \pm 0.10}$ \\
\bottomrule
\end{tabularx}
\caption{Quantitative test-set ablation results. LDM models were trained and tested on patient-level dataset, whereas CNet-based models used the lesion-level dataset. Values are mean $\pm$ SD across the three orthogonal planes. M0 is the pretrained MAISI-v2 model without fine-tuning. CNet, ControlNet; csPCa, clinically significant prostate cancer class conditioning; EMA, exponential moving average; FID, Fréchet inception distance; KID, kernel inception distance; LDM, latent diffusion model; Prec, precision; Rec, recall; relLoss, relative advantage loss; relWLoss, relative advantage weighted loss; wLoss, region-weighted loss.}
\end{table}

External fidelity of M7, computed with an equally sized synthetic set generated from PI-CAI test set masks ($n = 283$), was FID $3.83 \pm 0.51$, KID $0.0028 \pm 4.4\times 10^{-4}$, precision $0.41 \pm 0.15$, and recall $0.24 \pm 0.13$, all worse than in-house values.

\subsection{csPCa classification}
\begin{figure}[H]
\centering
\includegraphics[width=1.0\linewidth]{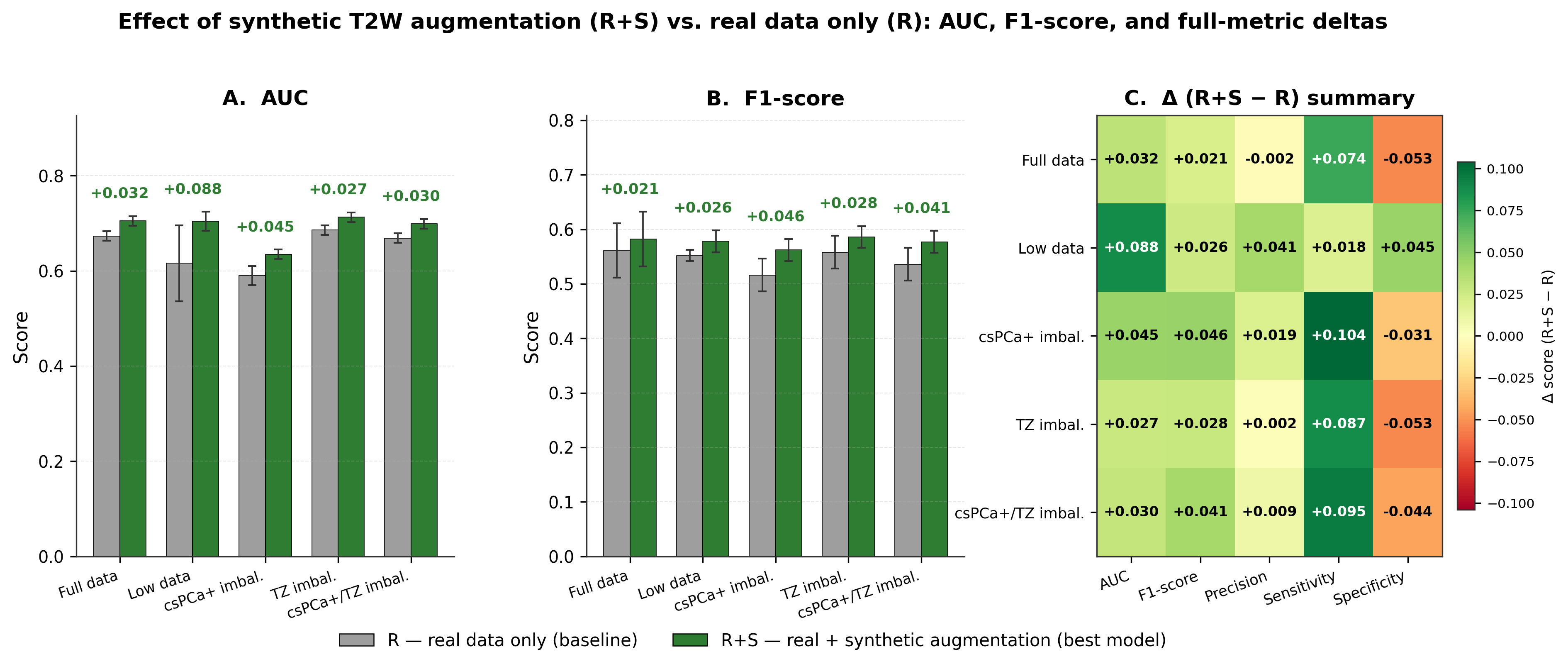}
\caption{Effect of synthetic T2-weighted (T2W) augmentation (R+S) versus real data only (R) in the five training scenarios: test AUC (A) and F1-score (B), shown as mean and SD across five seeds with the R+S $-$ R difference above each pair, and heatmap of the R+S $-$ R differences for all test metrics (C). AUC, area under the receiver operating characteristic curve; csPCa, clinically significant prostate cancer; imbal., imbalance; TZ, transition zone.}
\label{fig:classification_effects}
\end{figure}

In all five scenarios, the best validation classifiers using synthetic augmentation achieved higher test AUC than the real-only baselines (Table \ref{tab:classification_results}; Figure \ref{fig:classification_effects}). Mean AUC gains were $+0.032$ (95\% CI: $0.011$--$0.053$) with full training data, $+0.088$ ($0.049$--$0.126$) in the low-data regime, $+0.045$ ($0.007$--$0.083$) under csPCa+ imbalance, $+0.027$ ($0.007$--$0.046$) under TZ lesion underrepresentation, and $+0.030$ ($0.002$--$0.062$) under their combination, all significant after BH correction (adjusted $p = 0.032, 0.016, 0.036, 0.032$, and $0.043$). Sensitivity and F1-score increased in all scenarios, whereas specificity decreased in all but the low-data regime.

\begin{table}[H]
\centering
\label{tab:classification_results}
\vspace{0.5em}
\scriptsize
\begin{tabularx}{\textwidth}{l c c c c*{5}{C}}
\toprule
 & & \multicolumn{3}{c}{\textbf{Synthetic criteria}} & \multicolumn{5}{c}{\textbf{Test results}} \\
\cmidrule(lr){3-5} \cmidrule(lr){6-10}
\textbf{Scenario} & \textbf{Data} & \textbf{Sz} & \textbf{Prop} & \textbf{Sel} & \textbf{AUC} & \textbf{F1} & \textbf{Prec} & \textbf{Sens} & \textbf{Spec} \\
\midrule
\multirow{2}{*}{Full data} & R & -- & -- & -- & $0.673 \pm 0.01$ & $0.561 \pm 0.05$ & $0.487 \pm 0.02$ & $0.669 \pm 0.12$ & $0.576 \pm 0.06$ \\
 & R+S & 10 & TR-inv & Bound & $0.705 \pm 0.01$ & $0.582 \pm 0.05$ & $0.485 \pm 0.01$ & $0.743 \pm 0.15$ & $0.523 \pm 0.07$ \\
\cmidrule(lr){1-10}
\multirow{2}{*}{Low data} & R & -- & -- & -- & $0.616 \pm 0.08$ & $0.552 \pm 0.01$ & $0.452 \pm 0.04$ & $0.721 \pm 0.06$ & $0.457 \pm 0.13$ \\
 & R+S & 20 & TR-ratio & Bound & $0.704 \pm 0.02$ & $0.578 \pm 0.02$ & $0.493 \pm 0.06$ & $0.739 \pm 0.16$ & $0.502 \pm 0.26$ \\
\cmidrule(lr){1-10}
\multirow{2}{*}{csPCa+ imbal.} & R & -- & -- & -- & $0.590 \pm 0.02$ & $0.516 \pm 0.03$ & $0.414 \pm 0.02$ & $0.700 \pm 0.12$ & $0.394 \pm 0.15$ \\
 & R+S & 20 & TR-inv & Fixed & $0.635 \pm 0.01$ & $0.562 \pm 0.02$ & $0.433 \pm 0.01$ & $0.804 \pm 0.07$ & $0.363 \pm 0.07$ \\
\cmidrule(lr){1-10}
\multirow{2}{*}{TZ imbal.} & R & -- & -- & -- & $0.686 \pm 0.01$ & $0.558 \pm 0.03$ & $0.481 \pm 0.03$ & $0.669 \pm 0.07$ & $0.560 \pm 0.07$ \\
 & R+S & 10 & Balanced & Bound & $0.713 \pm 0.01$ & $0.586 \pm 0.02$ & $0.483 \pm 0.01$ & $0.756 \pm 0.10$ & $0.507 \pm 0.09$ \\
\cmidrule(lr){1-10}
\multirow{2}{*}{csPCa+/TZ imbal.} & R & -- & -- & -- & $0.669 \pm 0.01$ & $0.536 \pm 0.03$ & $0.479 \pm 0.03$ & $0.613 \pm 0.06$ & $0.594 \pm 0.06$ \\
 & R+S & 10 & TR-inv & Bound & $0.699 \pm 0.01$ & $0.577 \pm 0.02$ & $0.488 \pm 0.02$ & $0.708 \pm 0.05$ & $0.550 \pm 0.05$ \\
\bottomrule
\end{tabularx}
\caption{csPCa test-set results of the best configurations trained with real data only (R) and with synthetic augmentation (R+S) in each scenario. Synthetic criteria apply only to R+S: Sz, synthetic size (\% of the real training set); Prop, class proportion; Sel, case selection (Table \ref{tab:config}). Values are mean $\pm$ SD across five seeds. AUC, area under the curve; Bound, uncertainty-bounded selection; csPCa, clinically significant prostate cancer; TR-inv, inverse ratio; TR-ratio, real ratio; TZ, transition zone.}
\end{table}

Across scenarios, the best-performing configuration used synthetic augmentation equivalent to 10--20\% of the real training set size, with uncertainty-guided (Bounded) selection in four of five scenarios. Class proportion was scenario-dependent: TR-inv ratio was optimal in three scenarios (full data, csPCa+ imbalance, and csPCa+/TZ imbalance), whereas the low-data regime and TZ-imbalance scenarios favored the TR\_ratio and Balanced options, respectively.

In the cross-institutional evaluation (synthetic data from in-house masks added to PI-CAI real data), the cross-validation-selected configuration (20\% synthetic data, balanced class composition, uncertainty-bounded selection) achieved a test AUC of $0.654 \pm 0.02$, versus $0.641 \pm 0.02$ for real data alone (non-significant mean difference $+0.013$; 95\% CI: $-0.022$ to $0.048$; $p = 0.397$).

\section{Discussion}
Adapting MAISI-v2 into a mask- and class-conditioned T2W prostate MRI synthesis model improved image fidelity over the pretrained framework. In the cumulative ablation, fine-tuning on prostate data produced the largest FID reduction (13.28 to 3.59), indicating that domain adaptation accounted for most of the fidelity gain for this sequence. Class conditioning and EMA reduced FID further to 2.79, with EMA increasing recall at a small cost in precision, suggesting greater sample diversity. The ControlNet enabled spatial control over zonal anatomy and lesion location and, combined with relWLoss yielded best FID (2.04), precision and recall. Fidelity was lower on PI-CAI, likely owing to the domain gap between images generated with in-house acquisition characteristics and the multi-vendor, multicenter PI-CAI reference distribution.

The synthetic images also showed downstream utility. Compared with matched real-only T2W classifiers, synthetic augmentation significantly increased AUC in all five simulated scarcity and imbalance training data scenarios ($+0.027$ to $+0.088$), with configurations selected on validation and a held-out test set containing no cases used to train the generative model. The largest gain occurred in the low-data regime, possibly because synthetic cases represent a larger relative addition to a small training set. Gains were smaller under TZ lesion underrepresentation, alone and combined with csPCa-positive imbalance ($+0.027$ and $+0.030$), which may reflect the radiological overlap of TZ lesions with benign prostatic hyperplasia on T2W \cite{turkbey2019}. No single configuration was optimal across scenarios, although synthetic data consistently made up 10--20\% of the training set, and uncertainty-bounded selection and an inverted class ratio were selected in four and three of the five scenarios respectively. In the cross-institutional evaluation, adding in-house synthetic data to PI-CAI training data yielded a small, non-significant gain ($+0.013$). Given the limited statistical power, this is consistent with, but does not establish, a dependence of the benefit on the match between synthetic and target cohorts.

Comparison with previous diffusion-based T2W prostate MRI synthesis studies is limited. Per-plane FIDs of 5.27--8.09 \cite{bashkanov2024} and an ImageNet-based FID of 307.4 \cite{giardina2025} have been reported, but differing feature extractors, datasets, and sample sizes preclude direct comparison. For csPCa classification, T2W synthetic augmentation increased AUC by 0.04 (0.76 to 0.80) in an in-house cohort \cite{grabke2025a}. This gain lies within our range, although differences in task definition and cohort size limit further comparison. In \cite{grabke2025b}, the benefit of synthetic data was greatest at the smallest target-institution volumes, consistent with our largest gain in the low-data regime. However, its multi-sequence synthesis achieved stronger cross-institutional gains than our T2W-only transfer to PI-CAI. Our full-data real-only baseline (AUC $0.673 \pm 0.01$) was comparable to a reported T2W-only classifier (AUC 0.70) \cite{sun2025} and, as expected, below mpMRI-based models \cite{shinmoto2015}.

This study has limitations that point to future work. First, the generative model was trained on a single-center cohort, which probably contributed to the lower fidelity and non-significant classification gain on PI-CAI. Second, synthesis and classification were restricted to T2W, a deliberate choice because T2W is routinely acquired and requires no multi-sequence harmonization. However, it limits absolute performance, since DWI and ADC carry most diagnostic information in the peripheral zone and complement T2W in the TZ. Third, data constraints were simulated by subsampling a single cohort rather than observed in real low-resource or imbalanced settings. Fourth, synthetic augmentation was not compared with conventional augmentation or other non-generative strategies. Fifth, each R vs R+S classifier comparison relied on five paired runs, which, together with correction for multiple comparisons, limits statistical power. Sixth, the ablation was cumulative, assessing each component conditional on those preceding it. This assessment was done with 2.5D feature-based metrics on 130--160 volumes, without bootstrap confidence intervals or an expert reader study, and these metrics are sensitive to sample size and to the choice of feature extractor. Seventh, memorization of training images, relevant both to patient privacy and to interpreting downstream gains, was not assessed. Finally, hyperparameter exploration and sensitivity analyses of the generative and downstream classification models were limited, which may have constrained absolute performance.

In conclusion, the proposed model achieved higher T2W prostate MRI fidelity than pretrained MAISI-v2, and with the best-performing configuration, its synthetic images significantly improved T2W-only csPCa classification in all five simulated training data scenarios, including TZ lesion underrepresentation, alone and with severe class imbalance. The benefit depended on synthetic case selection and was not significant on an independent dataset. Synthetic T2W augmentation may therefore complement real data when lesion subtypes or classes are underrepresented, but its utility should be confirmed in the target cohort before clinical use.

\section*{Declarations}

\subsection*{Ethics approval and consent to participate}
The study was approved by the institutional Ethics Committee (PR(AG)02/2021). Data were collected and pseudonymized in accordance with the ethical committee's approval, with a single code per case. Data Protection Impact Assessment was completed and reviewed by the committee.

\subsection*{Consent for publication}
The Ethics Committee granted a waiver of written informed consent, as participants were verbally informed at the time of specimen collection about the research use of their data.

\subsection*{Availability of data and materials}
For data access, please contact the authors of the manuscript.

\subsection*{Competing interests}
XR-P, AV-R, II-R, CP-L and AJ-P are employed by Quibim and have no competing interests to declare that are relevant to the content of this article.

\subsection*{Funding}
The authors state that this work has not received any funding.

\subsection*{Authors' contributions}
Study conception and design: XR-P, AJ-P, AAR, JM. Acquisition of data: AAR, JM, XR-P, CP-L. Analysis and interpretation of data: XR-P, AV-R, II-R, CP-L. Drafting of manuscript: XR-P, AJ-P. Critical revision: all authors.

\subsection*{Acknowledgements}
Not applicable.


\end{document}